\documentclass[sigconf, screen, nonacm]{acmart}

\usepackage{microtype}
\usepackage{graphicx}
\usepackage{subcaption}
\usepackage{booktabs}
\usepackage{bm} 
\usepackage{multirow}
\usepackage{makecell}
\usepackage{tabularx} 
\usepackage{amsmath}
\usepackage{mathtools}
\usepackage{amsthm}
\usepackage{threeparttable}
\usepackage{hyperref}
\usepackage{amsfonts}

\begin{document}

\title{LBFTI: Layer-Based Facial Template Inversion for Identity-Preserving Fine-Grained Face Reconstruction}

\author{Zixuan Shen}
\affiliation{%
  \institution{College of Cyber Security, Jinan University}
  \city{Guangzhou}
  \state{Guangdong}
  \country{China}
}
\email{m15195901120@163.com}

\author{Zhihua Xia}
\authornote{The corresponding author.}
\affiliation{%
  \institution{College of Cyber Security, Jinan University}
  \city{Guangzhou}
  \state{Guangdong}
  \country{China}
}
\email{xia\_zhihua@163.com}

\author{Kaikai Gan}
\affiliation{%
  \institution{College of Cyber Security, Jinan University}
  \city{Guangzhou}
  \state{Guangdong}
  \country{China}
}
\email{kkgan406@163.com}

\author{Peipeng Yu}
\affiliation{%
  \institution{College of Cyber Security, Jinan University}
  \city{Guangzhou}
  \state{Guangdong}
  \country{China}
}
\email{ypp865@163.com}

\renewcommand{\shortauthors}{Shen et al.}

\begin{abstract}
  In face recognition systems, facial templates are widely adopted for identity authentication due to their compliance with the data minimization principle. However, facial template inversion technologies have posed a severe privacy leakage risk by enabling face reconstruction from templates. This paper proposes a Layer-Based Facial Template Inversion (LBFTI) method to reconstruct identity-preserving fine-grained face images. Our scheme decomposes face images into three layers: foreground layers (including eyebrows, eyes, nose, and mouth), midground layers (skin), and background layers (other parts). LBFTI leverages dedicated generators to produce these layers, adopting a rigorous three-stage training strategy: (1) independent refined generation of foreground and midground layers, (2) fusion of foreground and midground layers with template secondary injection to produce complete panoramic face images with background layers, and (3) joint fine-tuning of all modules to optimize inter-layer coordination and identity consistency. Experiments demonstrate that our LBFTI not only outperforms state-of-the-art methods in machine authentication performance, with a 25.3\% improvement in TAR, but also achieves better similarity in human perception, as validated by both quantitative metrics and a questionnaire survey.

\end{abstract}

\keywords{Facial template inversion, face reconstruction, facial attribute, layer-based network, identity preservation}

\maketitle

\section{Introduction}
Leveraging advantages such as non-contact operation and high convenience, face recognition (FR) systems have been widely deployed in scenarios \cite{hasan2023presentation} including mobile payment, airport customs clearance, and office access control. Early FR systems directly stored raw face images for identity verification \cite{bowyer2004face}. However, such images contain substantial redundant information irrelevant to recognition tasks, posing significant privacy and security risks. Moreover, Article 5(1)(c) of the \textit{European Union's General Data Protection Regulation (GDPR)} explicitly stipulates that ``Personal data shall be adequate, relevant and limited to what is necessary in relation to the purposes for which they are processed (`data minimisation')'' \cite{gdpr2016}. Similarly, Article 12 of China's \textit{Measures for the Security Administration of the Application of Face Recognition Technology} mandates that ``Except as required by law or with the separate consent of the natural person, no organization or individual may store raw face images, pictures, or videos, and shall adopt information with distinct features for identity recognition'' \cite{ChinaFaceRecognitionMeasures2025}. Therefore, as a compact and discriminative representation of raw face images, facial templates have emerged as the core storage object in mainstream FR systems.

\begin{figure}[t]
  \centerline{\includegraphics[width=0.48\textwidth]{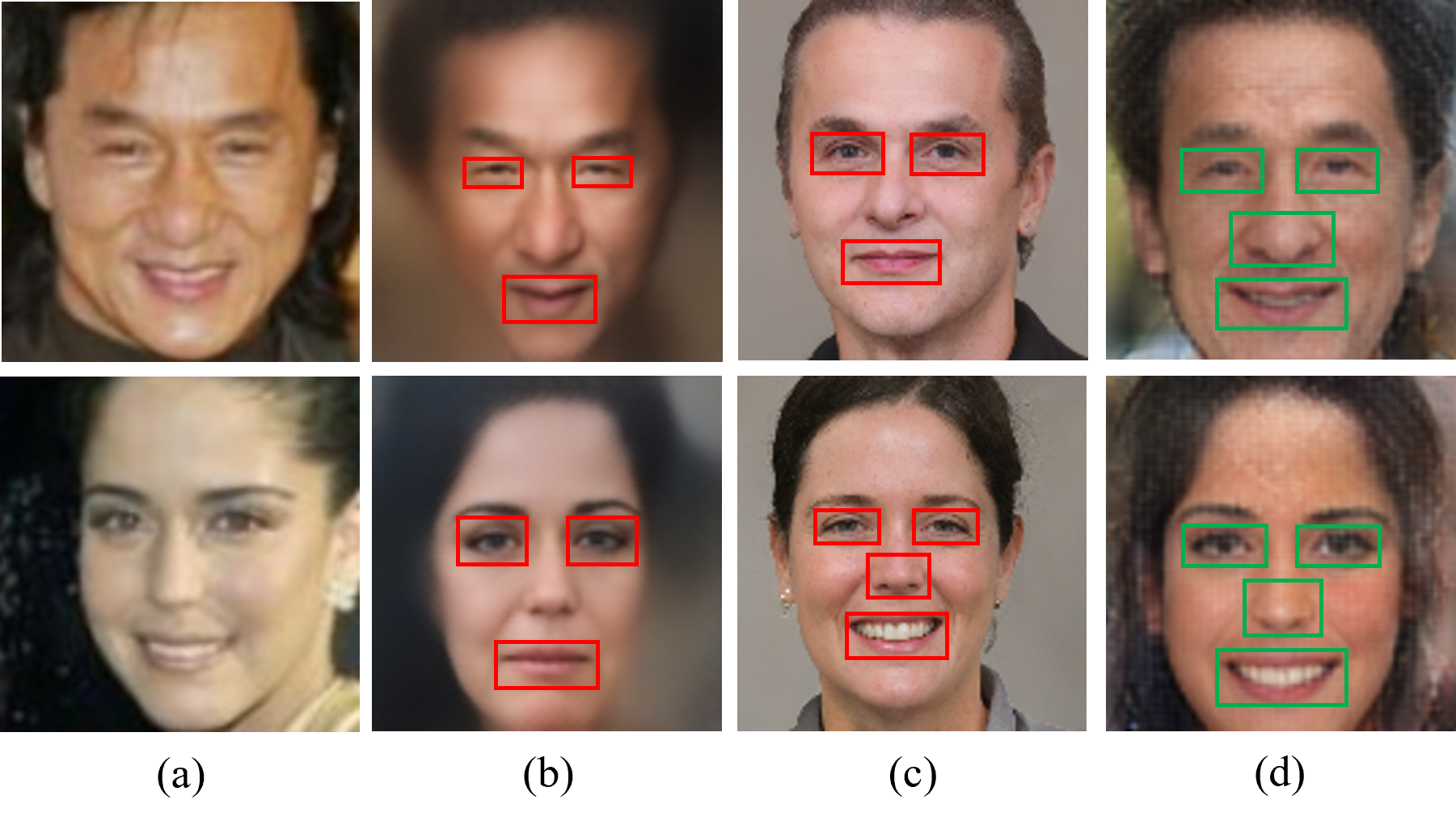}}
  \caption{Limitations of existing FTI methods: (a) original images, (b) images reconstructed by a custom-built inverter \cite{shahreza2024vulnerability}, and (c) images reconstructed by a pre-trained network-based inverter \cite{dai2026clipfti}. Although the images in columns (b) and (c) can pass machine authentication, their facial attributes are inconsistent with those in column (a) from the perspective of human visual perception. In contrast, the images in (d) reconstructed by our proposed LBFTI can not only pass machine authentication, but also maintain consistent facial attributes with the original images.}

  \label{fig:FTI}
\end{figure}

However, storing facial templates is far from fully secure. Since templates encode critical identity information, they can be reconstructed into face images via Facial Template Inversion (FTI) techniques \cite{deng2019arcface,boutros2022elasticface}, which also exposes security vulnerabilities in practical FR systems. As shown in Fig. \ref{fig:FTI}, existing FTI methods only optimize machine authentication performance while neglecting human visual consistency. Their reconstructed images exhibit obvious visual deviations and inconsistent facial attributes from the originals. Since real-world attacks require passing both machine verification and manual inspection, such low-fidelity results are impractical for real scenarios. To address this issue, we propose a novel Layer-Based Facial Template Inversion (LBFTI) method. With a layer-structured network and a rigorous three-stage training strategy, LBFTI achieves high-precision facial attribute reconstruction. The main contributions of this work are summarized as follows:

\begin{itemize}
    \item \textbf{Innovative layer-based architecture}. We first propose a ``foreground-midground-background'' layer-based inversion framework. It defines the facial attributes critical for human visual identity perception  (eyebrows, eyes, nose, and mouth) as the foreground, the skin region as the midground, and the remaining parts as the background. Leveraging a rigorous three-stage training strategy tailored for the four dedicated foreground generators, midground generator, and panorama generator, LBFTI significantly enhances the human-perceptual authenticity of facial attributes in the reconstructed face images.

    \item \textbf{Human-perception-oriented evaluation metrics}. Humans primarily judge the identity by observing facial features. Accordingly, we design two dedicated metrics for fine-grained facial attribute assessment: Facial Attribute Pixel Deviation, which quantifies pixel-level discrepancies in foreground layers, and Facial Attribute Perceptual Consistency, which evaluates their deep semantic consistency. In addition to these objective metrics, we conduct a subjective questionnaire survey where participants visually match generated images with original ones. This combination fills the gap in traditional evaluation protocols that only focus on machine authentication and overlook human-perceptual realism.

    \item \textbf{Superior performance against SOTA methods}. We conduct extensive experiments on the LFW benchmark dataset, comparing LBFTI with three representative SOTA methods: Adv-StyleGAN3 (NeurIPS 2023) \cite{otroshi2023face}, MSE-StyleGAN3 (TBIOM 2024) \cite{shahreza2024template}, and CLIP-FTI (AAAI 2026) \cite{dai2026clipfti}. Specifically, in Type-II attack scenarios, LBFTI achieves a 25.3\% improvement in TAR compared with these SOTA methods, while achieving relative reductions of 33.5\% in FAPD and 2.9\% in FAPC. In the subjective questionnaire survey, 85\% of participants voted that images reconstructed by LBFTI are more similar to the original faces. These results validate its advantages in both machine authentication accuracy and human-perceptual realism.

\end{itemize}

\section{Related Work}
Research on facial template inversion (FTI) can be classified into three types based on the type of feature representation for inversion: traditional handcrafted feature inversion \cite{adler2003sample,mignon2013reconstructing}, confidence vector inversion \cite{fredrikson2015model}, and deep facial feature inversion \cite{mai2018reconstruction}. This paper focuses on the third type noted above, as the facial templates processed in face recognition (FR) systems are derived from deep facial features. Relevant works in this field are further divided into two categories, as summarized in Table \ref{tab:FTI_studies}.

\begin{table*}[t]
  \caption{Related research on facial template inversion.}
  \label{tab:FTI_studies}
  \centering
  \scriptsize
  \begin{tabular}{@{}llllll@{}}
    \toprule
    References & Types & Inverter Backbones & Target Models & Evaluation Metrics & Code Available \\
    \midrule
    \cite{mai2018reconstruction} & \multirow{4}{*}{Custom-Built} & NbNet (Deconv+Conv) & FaceNet & TAR/FAR & \checkmark \\
    \cite{duong2020vec2face} &  & FCG & ArcFace & MS-SSIM, IS, Matching Acc & $\times$ \\
    \cite{shahreza2022face} &  & Deconv+Conv & ArcFace, ElasticFace & TMR/FMR & \checkmark \\
    \cite{shahreza2024vulnerability} &  & DSCasConv (Deconv+Conv) & ArcFace & TMR/FMR & \checkmark \\
    \midrule
    \cite{dong2021towards} & \multirow{10}{*}{\makecell[l]{Pre-trained \\network-based}} & StyleGAN2 & InsightFace & TAR/FAR & $\times$ \\
    \cite{vendrow2021realistic} &  & StyleGAN2 & FaceNet & L2 Distance, Cosine Distance & \checkmark \\
    \cite{dong2023reconstruct} &  & StyleGAN2 & ArcFace & TAR/FAR & \checkmark \\
    \cite{shahreza2023inversion} &  & StyleGAN3 & ArcFace, ElasticFace & TMR/FMR & \checkmark \\
    \cite{otroshi2023face} &  & StyleGAN3 & ArcFace, ElasticFace & TMR/FMR & \checkmark \\
    \cite{shahreza2023template} &  & GNeRF (StyleGAN2) & ArcFace, ElasticFace & TMR/FMR & \checkmark \\
    \cite{huang2023comprehensive} &  & GNeRF (StyleGAN2) & ArcFace, ElasticFace & TMR/FMR & \checkmark \\
    \cite{shahreza2024template} &  & StyleGAN3 & ArcFace, ElasticFace & TMR/FMR & \checkmark \\
    \cite{Shahreza_2025_CVPR} &  & Stable Diffusion & ArcFace, ElasticFace & TMR/FMR & $\times$ \\
    \cite{dai2026clipfti} &  & CLIP-FTI (StyleGAN3) & ArcFace, ElasticFace & TAR/FAR, LPIPS, FAMSE & \checkmark \\
    \midrule
    Our LBFTI & Custom-Built & Our LBFTI (Upsample+Conv) & ArcFace, ElasticFace & TAR/FAR, FFPD, FFPC & \checkmark \\
    \bottomrule
  \end{tabular}
\end{table*}

\textbf{FTI based on custom-built networks}. Mai et al. \cite{mai2018reconstruction} first systematically studied the inversion security of deep facial features (i.e., facial templates) , proposing the neighborly de-convolutional network (NbNet). NbNet consists of multiple Neighborly Deconvolution Blocks (NbBlocks), where each block integrates one deconvolution operation and multiple convolution operations. It achieves feature refinement through intra-block channel concatenation, thereby enhancing the detail richness of reconstructed face images. Shahreza et al. \cite{shahreza2022face} improved NbNet by designing an inverter with six blocks, each consisting of one deconvolution and three convolution operations, and fusing intra-block features using skip connections. Later, Shahreza et al. \cite{shahreza2024vulnerability} further introduced the DSCasConv block, which introduces dense skip connections between convolution operations within each block and generates the final image via four parallel convolution layers with different kernel sizes, significantly improving the identity authentication pass rate. In addition, Duong et al. \cite{duong2020vec2face} proposed the Feature-Conditioned Generator (FCG), which integrates facial templates as conditional inputs into multi-scale feature maps of the generator. They also designed a bijection metric loss to directly utilize identity distribution in the image domain, and optimized the model by combining bijection distillation loss, adversarial loss, and pixel loss.

\textbf{FTI based on pre-trained networks}. A large body of existing research adopted pre-trained StyleGAN-series generative networks \cite{karras2020analyzing,karras2021alias} as the backbone for FTI inverters, with research efforts primarily focused on establishing effective template-to-latent-code mappings and refining latent code search strategies. Dong et al. \cite{dong2021towards} first applied StyleGAN2 to FTI, constructing a mapping between facial templates and StyleGAN2 input latent codes by training a regression model. In their follow-up work \cite{dong2023reconstruct} incorporated the Genetic Algorithm (GA) to optimize latent code selection, thereby further enhancing reconstruction accuracy. Similarly, Vendrow et al. \cite{vendrow2021realistic} leveraged the Simulated Annealing Algorithm (SAA) to search for optimal latent codes, effectively alleviating the overfitting issue of regression-based mapping models. A series of works by Shahreza et al. \cite{shahreza2023inversion,otroshi2023face,shahreza2024template} adopted pre-trained StyleGAN3 as the generation backbone. They designed a mapping network to convert facial templates into the StyleGAN3 intermediate latent codes, and optimized the model by combining latent code discriminator loss, pixel loss, and ID loss. Shahreza et al. \cite{shahreza2023template,huang2023comprehensive} further extended this framework to 3D face reconstruction, employing the pre-trained GNeRF generator, which embeds StyleGAN2 as its core component. In addition, Dai et al. \cite{dai2026clipfti} proposed CLIP-FTI, an enhanced method that integrates an attribute alignment adapter based on StyleGAN3. Specifically, it first converts facial templates into CLIP-based semantic attribute embeddings, then trains a joint mapping network using both facial templates and CLIP semantic embeddings to map them into the latent codes of StyleGAN3, enhancing the attribute consistency of reconstructed images. Furthermore, Shahreza et al. \cite{Shahreza_2025_CVPR} proposed an adapter module to project leaked templates into the feature embedding space of Arc2Face, and then injected them into the UNet decoder of Stable Diffusion to achieve face image reconstruction.

While the two aforementioned categories of methods have achieved phased advancements in boosting machine-based identity verification pass rates, the reconstructed face images still have severe deficiencies in human visual perception of identity. Specifically, key facial attributes (e.g., eyebrows, eyes, nose, and mouth) exhibit considerable discrepancies compared with the original images. Such inconsistencies directly render the reconstructed images unable to pass manual identity verification, making FTI attacks completely invalid in practical scenarios with human review. Meanwhile, existing methods primarily focus on machine authentication metrics (e.g., TAR/FAR), lacking dedicated quantitative metrics for human-perceptual evaluation. To address these limitations, this paper proposes a novel Layer-Based Facial Template Inversion (LBFTI) method, which achieves high-precision reconstruction of facial attributes and enhances identity consistency in human visual perception. Furthermore, this paper introduces two novel quantitative human-perception-oriented evaluation metrics (FAPD and FAPC) to bridge the gap in existing evaluation frameworks.

\section{System Model}
\label{sec:CommGen15_dataset}

\subsection{Model Definition}

In face recognition (FR) systems, facial template inversion (FTI) can be formally defined by the following mathematical formulation:

\textbf{Face image set}. Let $\mathcal{X}$ denote the set of all legitimate face images. Each face image $x \in \mathcal{X}$ is an $H \times W \times 3$ tensor, corresponding to a color face image with height $H$ and width $W$. In our experiments, we set $H=W=128$.

\textbf{Facial template set}. Let $\mathcal{T}$ denote the set of all facial templates extracted from $\mathcal{X}$. Each facial template $t \in \mathcal{T}$ is a $d$-dimensional real-valued feature vector. Specifically, we use 512-dimensional facial templates extracted by ArcFace-IResNet100 \cite{deng2019arcface} and ElasticFace-IResNet100 \cite{boutros2022elasticface} in our experiments.

\textbf{Facial template extractor}. Define $E: \mathcal{X} \to \mathcal{T}$ as the facial template extraction function of the target FR system. It takes a face image $x$ as input and outputs the corresponding template $t=E(x)$. In attack scenarios, $E$ is regarded as a black box: the attacker has no access to its network architecture, parameters, or training details, but can obtain the corresponding template by inputting images into $E$.

\textbf{Facial template inverter}. Define $G:\mathcal{T} \to \mathcal{X}$ as the FTI function, which is instantiated by a generative neural network $g_{\theta}$, where $\theta$ denotes the trainable parameters. It takes a facial template $t$ as input and outputs a reconstructed face image $\hat{x}=G(t)$. The core objective of the inverter is to maximize identity consistency and visual similarity between $\hat{x}$ and the original image $x$, where $x$ satisfies $t=E(x)$.

\begin{figure*}[t]
  \centerline{\includegraphics[width=1.0\textwidth]{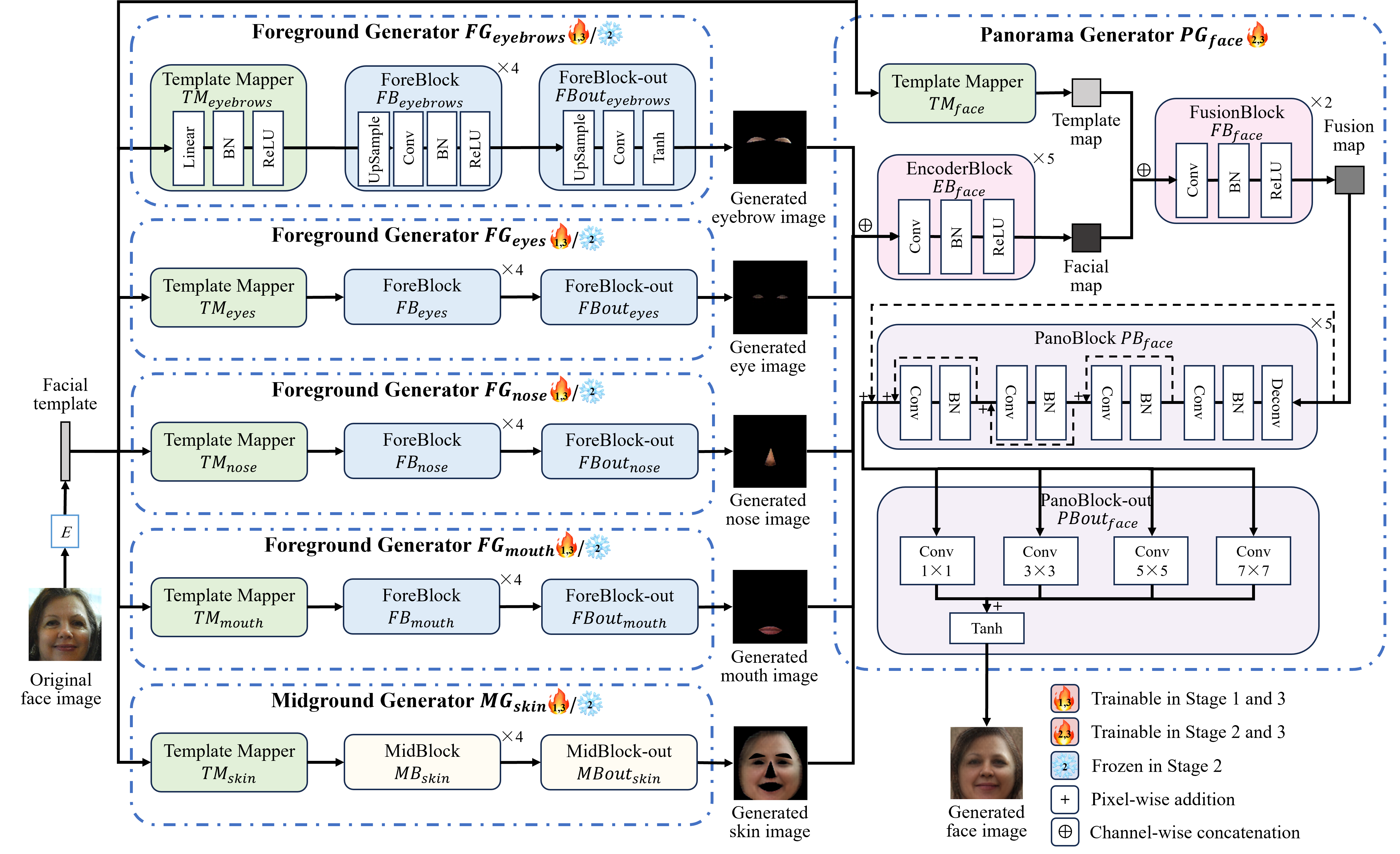}}
  \caption{Architecture of the proposed LBFTI. It comprises four foreground generators, one midground generator, and one panorama generator, governed by a three-stage training strategy: Stage 1 trains the foreground and midground generators; Stage 2 freezes the former two and trains the panorama generator; Stage 3 jointly fine-tunes all modules.}

  \label{fig:LBFTI}
\end{figure*}

\subsection{Attacker's Capabilities}

In line with real-world attack scenarios, we make reasonable assumptions regarding the attacker's capabilities as follows:

\textbf{Obtaining target templates}. The attacker can acquire the target subject's facial template $t^*=E(x^*)$, where $x^*$ denotes the target subject's original face image, via legitimate or malicious means, such as database leakage, internal data breaches, or transmission interception.

\textbf{Black-box access to the target model}. The attacker is capable of inputting arbitrary face images into the target FR system to retrieve the corresponding facial templates, thus constructing a large-scale training dataset $\{(x,t)\}$ where $t=E(x)$ for training the inverter $G$. However, the attacker has no access to the internal details of the template extractor $E$, including its network architecture, parameter configurations, and loss function design, nor can they leverage white-box information (e.g., gradient feedback) to optimize the inverter.

\subsection{Attacker's Objectives}

Given the target template $t^*=E(x^*)$, the attacker's core objective is to train a high-performance inverter $G$ to ensure that the reconstructed image $\hat{x}=G(t^*)$ satisfies the following three requirements:

\textbf{Passing target model authentication}. When inputting $\hat{x}$ into the target template extractor to extract the facial template $\hat{t}=E(\hat{x}$), the similarity between $\hat{t}$ and $t^*$ must exceed the system's verification threshold, thus passing the identity verification process.

\textbf{Cross-model generalization capability}. When inputting $\hat{x}$ into an ``unseen model'' that was not included in the inverter’s training pipeline to extract the template $\hat{t}=E(\hat{x})$, the similarity between $\hat{t}$ and $t^*$ must exceed the verification threshold of this unseen model.

\textbf{Human visual consistency}. Human primarily judge identity by observing facial features. Thus, $\hat{x}$ should preserve key facial attributes of $x^*$, including eyebrow texture, eye contour, nose structure, and mouth shape. This guarantees consistent identity perception from a human visual perspective.

\section{Methodology}
\label{sec:methodology}

To achieve fine-grained facial reconstruction, our LBFTI framework independently models distinct facial component layers and fuses them into a complete face image. This layered design addresses the limitation of conventional monolithic generative models, which struggle to reconstruct precise facial details. Our framework adopts four foreground generators, one midground generator, and one panorama generator, optimized via a three-stage training strategy (see Fig. \ref{fig:LBFTI}). In this section, we elaborate the core modules, loss functions, and training strategy of LBFTI.

\subsection{Core Modules}

Our proposed LBFTI framework consists of three core modules: four foreground generators, one midground generator, and one panorama generator. The detailed design and functionalities of these modules are elaborated below.

\textbf{Foreground generator}. Each foreground generator is dedicated to generating a single fine-grained facial attribute component: eyebrows, eyes, nose, and mouth. Since these regions are localized and structurally simple, we adopt a lightweight network architecture consisting of one template mapper, four ForeBlocks, and one ForeBlock-out. This lightweight design keeps the foreground generation pipeline concise, while performance gains are mainly attributed to our disentangled modeling strategy. The generator upsamples $512$-dimensional facial templates to generate $128\times128$ color foreground layers corresponding to key facial attributes. The template mapper transforms facial templates into compatible feature representations for downstream ForeBlocks, consisting of a linear layer, a BatchNorm layer, and a ReLU activation layer. Each ForeBlock conducts sequential upsampling and feature refinement using an upsampling layer,  a convolution layer, a BatchNorm layer, and a ReLU activation layer. The ForeBlock-out generates normalized color foreground layers via an upsampling layer, a convolution layer, and a Tanh activation layer.

\textbf{Midground generator}. Similar to the foreground generators, the midground generator only reconstructs the facial skin region. This region features continuous and simple textures and requires no complex feature learning. Accordingly, it features an identical lightweight architecture to the foreground generators, containing a template mapper, four MidBlocks, and a MidBlock-out with consistent layer configurations. The midground generator is designed to upsample $512$-dimensional facial templates into a $128\times128$ color midground layer corresponding to facial skin regions.

\textbf{Panorama generator}. Unlike the foreground and midground generators, the panorama generator integrates all foreground attribute layers, the midground skin layer, and background context to synthesize a complete and high-fidelity face image. This task is considerably more complex. It requires richer feature representations and advanced feature propagation mechanisms. Accordingly, the panorama generator adopts a more elaborate architecture with increased parameter capacity. Therefore, it adopts a more complex and elaborate design with larger parameter counts, consisting of one template mapper, five EncoderBlocks, two FusionBlocks, five PanoBlocks, and one PanoBlock-out. Specifically, the $128\times128\times3$ layers output from the foreground and midground generators are encoded into $4\times4$ facial maps via five EncoderBlocks. These maps then undergo fusion with template maps through FusionBlocks to enhance identity consistency. Finally, the fusion maps are fed into five PanoBlocks and one PanoBlock-out for progressive refinement, generating the final complete face image. Each EncoderBlock and FusionBlock comprises a  convolution layer, a BatchNorm layer, and a ReLU activation layer. Each PanoBlock integrates a deconvolution layer, four BatchNorm layers, and four convolution layers. Its design is inspired by the skip connection mechanism in ResNet \cite{he2016deep}, which facilitates the propagation of intra-module and inter-module features to alleviate feature forgetting. The PanoBlock-out is devised based on the multi-scale convolution paradigm of Inception \cite{szegedy2015going}, employing $1\times1$, $3\times3$, $5\times5$, and $7\times7$ convolution kernels to capture multi-scale contextual features of the feature maps. Ultimately, the final face image is normalized and output through a Tanh activation layer.

\subsection{Loss Functions}
To guarantee both identity consistency and visual similarity of the reconstructed images, LBFTI employs the following four complementary loss functions for model training:

\textbf{Template loss}. The template loss is designed to enforce identity consistency between the reconstructed image and the original image, ensuring that the facial template extracted from the reconstructed image is as close as possible to the original template. It quantifies the element-wise difference between the two templates using the mean squared error (MSE), with the formula defined as:
\begin{equation}
\label{eq:template_loss}
\mathcal{L}_{\text{tmp}} = \frac{1}{d} \sum_{i=1}^{d} (t_i - \hat{t}_i)^2,
\end{equation}
where $t_i$ and $\hat{t}_i$ denote the values of the $i$-th dimension of the facial template $t$ and the template $\hat{t}$ extracted from the reconstructed image, respectively, and $d$ denotes the dimensionality of the facial template.

\textbf{Pixel loss}. The pixel loss ensures pixel-level similarity between the reconstructed image and the original image by computing the MSE of corresponding pixel values across all RGB channels. Its mathematical expression is:
\begin{equation}
\label{eq:pixel_loss}
\mathcal{L}_{\text{pix}} = \frac{1}{H \times W \times 3} \sum_{i=1}^{H \times W \times 3} (x_i - \hat{x}_i)^2,
\end{equation}
where $x_i$ and $\hat{x}_i$ denote the $i$-th pixel values of the original image $x$ and the reconstructed image $\hat{x}$, respectively, and $H$ and $W$ denote the height and width of the input image. 

\textbf{Perceptual loss}. Pixel loss only captures low-level pixel differences and fails to capture high-level semantic features, such as texture and shape. To address this limitation, perceptual loss computes the feature-level difference between the original and reconstructed images using a pre-trained convolutional neural network. The formula is defined as:
\begin{equation}
\label{eq:perceptual_loss}
\mathcal{L}_{\text{per}} = \frac{1}{H_l \times W_l \times C_l} \sum_{l \in L} \left\| \phi_l(x) - \phi_l(\hat{x}) \right\|_2^2,
\end{equation}
where $\phi_l(\cdot)$ denotes the feature extraction function of the $l$-th layer of the pre-trained VGG-16 network \cite{zhang2018unreasonable}, and $H_l$, $W_l$, $C_l$ denote the height, width, and channel number of the feature map output by the $l$-th layer, respectively. Following
the recommendation in \cite{zhang2018unreasonable}, we select $l \in L = \{\text{ReLU1\_2, ReLU2\_2, ReLU3\_3, ReLU4\_3, ReLU5\_3}\}$.

\textbf{Attribute loss}. To retain fine-grained facial attributes in the reconstructed image, we introduce an attribute loss that measures the L1 norm discrepancy between attribute vectors predicted by a pre-trained classifier $A(\cdot)$. Built on ResNet-18 \cite{he2016deep} and pre-trained on CelebA dataset \cite{liu2015deep}, $A(\cdot)$ outputs a 40-dimensional vector. The attribute loss is defined as:
\begin{equation}
\label{eq:attribute_loss}
\mathcal{L}_{\text{att}} = \frac{1}{40} \sum_{i=1}^{40} \left| A(x)_i - A(\hat{x})_i \right|,
\end{equation}
where $A(x)_i$ and $A(\hat{x})_i$ denote the $i$-th attribute probability, ranging from 0 to 1 and indicating the presence of a specific attribute, for the original and reconstructed images respectively.

\subsection{Training Strategy}
\label{subsec:training_strategy}
To prioritize fine-grained facial attribute reconstruction prior to complete face reconstruction, our LBFTI employs a three-stage training strategy, detailed as follows:

\textbf{Stage 1}. We train four foreground generators $FG_{eyebrows}$, $FG_{eyes}$, $FG_{nose}$, $FG_{mouth}$ and one midground generator $MG_{skin}$ to achieve independent fine-grained reconstruction of eyebrows, eyes, nose, mouth, and skin respectively. Each generator is optimized using a multi-loss objective consisting of template loss, pixel loss, and perceptual loss, ensuring that the reconstructed components align with the structural constraints of facial templates while preserving texture details.

\textbf{Stage 2}. We first freeze the parameters of all generators trained in Stage 1 to retain their component-level reconstruction capabilities, then exclusively train the panorama generator $PG_{face}$. This module fuses the foreground and midground layers with secondary facial template injection to generate panoramic face images integrated with background layers. In addition to the template, pixel, and perceptual losses adopted in Stage 1, $PG_{face}$ is additionally regularized by an attribute loss to enforce consistency between the reconstructed and the target facial attributes

\textbf{Stage 3}. We unlock all generator parameters and perform end-to-end joint fine-tuning on the entire LBFTI framework. The goal of this stage is to enhance inter-layer feature coherence and the global identity consistency of the reconstructed face images. In this stage, each generator is optimized using the identical loss function configuration as it was in its respective training stage.

\section{Experiment}

\subsection{Experimental Setup}

\textbf{Experimental environment}. All experiments are implemented with the PyTorch framework, and the hardware platform is a server configured with a single 32GB NVIDIA V100 GPU. For all training phases, the batch size is set to $32$, and the resolution of all face images is fixed at $128\times128$ for consistency.

\textbf{Training strategy}. LBFTI follows the three-stage training strategy detailed in Section 4.1, with the specific hyperparameter settings as follows: 1) Stage 1 trains the foreground generators and midground generator for 100 epochs with a learning rate of 0.0002. 2) Stage 2 freezes the parameters of all modules trained in Stage 1, then trains the panorama generator for 100 epochs with a learning rate of 0.0002. 3) Stage 3 performs joint fine-tuning on the parameters of all model modules for 20 epochs with a learning rate of 0.0001.

\textbf{Datasets}. Three widely used benchmark datasets are employed in our experiments, with a clear division of training and test sets to ensure the validity of the experimental results: FFHQ \cite{karras2019style}, LFW \cite{learned2016labeled}, and AgeDB \cite{moschoglou2017agedb}. Specifically, FFHQ (Flickr-Faces High-Quality) is utilized as the training dataset, comprising 70,000 high-quality face images without identity annotations. LFW (Labeled Faces in the Wild) serves as the primary test dataset, which consists of 13,233 face images from 5,749 distinct individuals and encompasses significant variations in pose, facial expression, ethnicity, illumination, focus, and image resolution. AgeDB is adopted as the challenging test dataset, containing 16,488 images of 568 famous individuals with substantial variations in age, illumination, and pose, thus posing more stringent challenges for FTI than LFW.

\subsection{Evaluation Metrics}
 
To comprehensively evaluate LBFTI, we adopt the mainstream BLUFR protocol \cite{liao2014benchmark} to measure the machine authentication performance of reconstructed faces. Considering the lack of human-perception-oriented metrics in existing FTI research and inspired by psychological findings \cite{young1985matching,tanaka1993parts} that human recognition relies primarily on internal facial features (e.g., eyes, nose and mouth) rather than external attributes such as hair and facial contour, we propose two novel quantitative indicators focusing on core facial components to assess visual authenticity at both pixel and semantic levels. Furthermore, we conduct a questionnaire for subjective human evaluation. Detailed definitions and calculations of all metrics are presented below.

\textbf{Verification precision}. The BLUFR protocol takes True Accept Rate (TAR) and False Accept Rate (FAR) as core evaluation metrics for machine identity authentication. TAR is defined as the proportion of reconstructed images that pass the identity verification of the target subject at a fixed FAR constraint. Specifically, TAR is computed when the similarity between the reconstructed template and the target template exceeds the system’s predefined decision threshold at the given FAR. A higher TAR value indicates superior machine authentication performance. In the experiments, we report two types of TAR values (Type-I and Type-II) at FAR thresholds of 1\% and 0.1\% for a comprehensive evaluation: Type-I TAR refers to the verification result between reconstructed images and the original template-corresponding images of the target subject, while Type-II TAR refers to the verification result between reconstructed images and other images of the same target subject.

\textbf{FAPD}. We propose Facial Attribute Pixel Deviation (FAPD) to quantify the pixel-wise differences in the foreground layers (eyebrows, eyes, nose, and mouth), where these facial attributes are precisely localized via facial landmark detection \cite{kazemi2014one}. The calculation formula is defined as:
\begin{equation}
\label{eq:fapd}
\mathrm{FAPD}(x, \hat{x}) = \frac{1}{n} \sum_{i=1}^{n} (x_i - \hat{x}_i)^2,
\end{equation}
where $n$ denotes the total number of pixels in the foreground regions, and $x_i$ and $\hat{x}_i$ denote the $i$-th pixel values in the foreground regions of the original image $x$ and the reconstructed image $\hat{x}$, respectively.

\textbf{FAPC}. To quantify the deep semantic feature differences in the foreground layers, we further design Facial Attribute Perceptual Consistency (FAPC) for evaluation. The calculation framework of FAPC is consistent with the perceptual loss defined in Section 4.3, with the only difference that feature extraction via the pre-trained VGG-16 network is exclusively performed on the foreground layers.

\textbf{Questionnaire Survey}. We design an online questionnaire\footnote{\url{https://ks.wjx.com/vm/rCyczwU.aspx\#}} for human visual recognition evaluation. Participants are asked (1) to match the given reconstructed images to their corresponding original images and (2) to select the most similar reconstruction for each given original image. The detailed experimental settings and results are provided in the \textbf{S.M.} A.

\subsection{Comparative Experiments}
\label{sec:compare}
To demonstrate the superiority of our proposed LBFTI, we perform comparative experiments with three representative SOTA FTI methods covering different technical paradigms: Adv-StyleGAN3 \cite{otroshi2023face}, MSE-StyleGAN3 \cite{shahreza2024template}, and CLIP-FTI \cite{dai2026clipfti}. Adv-StyleGAN3 leverages adversarial training on real FFHQ data to realize template-to-latent inversion. MSE-StyleGAN3 adopts synthetic training data and MSE regression for latent alignment optimization. CLIP-FTI introduces an extra CLIP semantic module to enrich latent inversion with attribute-aware features.

\textbf{Performance on machine authentication}. Tables \ref{tab:tar_lfw} and \ref{tab:tar_agedb} show the TARs on the LFW and AgeDB datasets, respectively, to evaluate the machine recognition rates on the target models and unseen models. It can be seen that on the LFW dataset, when ArcFace is used as the target model, the Type-II TAR of LBFTI at 0.1\% FAR reaches 0.9509, which is about 30\% higher than that of SOTA methods; when ElasticFace is used as the unseen model, the Type-II TAR of LBFTI at 0.1\% FAR reaches 0.9408, which is about 60\% higher than that of SOTA methods, proving that the identity features of the generated images have stronger generalisability. On the AgeDB dataset with higher inversion difficulty, LBFTI still maintains an advantage, with the Type-II TAR at 0.1\% FAR improved by about 10\% compared to the SOTA methods, verifying the superiority of our LBFTI. Furthermore, extensive cross-model experiments adopting MagFace \cite{meng2021magface}, FaceNet \cite{schroff2015facenet}, AdaFace \cite{kim2022adaface}, SphereFace \cite{liu2017sphereface}, and CurricularFace \cite{huang2020curricularface} as unseen models further demonstrate the remarkable superiority and strong generalization capability of LBFTI. Detailed full results are provided in \textbf{S.M.} B.

\begin{table}[h]
    \centering
    \small 
    \caption{TARs on the LFW dataset.}
    \label{tab:tar_lfw}
    \resizebox{1.0\linewidth}{!}{
    \begin{tabular}{@{}c|c|c|*{3}{>{\centering\arraybackslash}p{1.1cm}|}>{\centering\arraybackslash}p{1.1cm}@{}}
        \toprule
        \textbf{\multirow{2}{*}{\makecell{Databases \\and Losses}}} & \textbf{\multirow{2}{*}{\makecell{Target Models/ \\Unseen Models}}} & \textbf{Inverters} & \multicolumn{2}{c|}{\textbf{TAR$\uparrow$ at 1\%FAR}} & \multicolumn{2}{c@{}}{\textbf{TAR$\uparrow$ at 0.1\%FAR}} \\
        \cmidrule(lr){4-5} \cmidrule(lr){6-7}
        & & & \textbf{Type-I} & \textbf{Type-II} & \textbf{Type-I} & \textbf{Type-II} \\
        \midrule
        \multirow{8}{*}{ArcFace}
        & \multirow{4}{*}{ArcFace} & Adv-StyleGAN3 & \textbf{1.0000} & 0.9294 & 0.9980 & 0.6980 \\
        & & MSE-StyleGAN3 & 0.9992 & 0.8800 & 0.9879 & 0.6602\\
        & & CLIP-FTI & 0.9998 & 0.8915& 0.9922& 0.6276\\
        & & LBFTI & 0.9998 & \textbf{0.9861} & \textbf{0.9992} & \textbf{0.9509} \\
        \cmidrule(lr){2-7}
        & \multirow{4}{*}{ElasticFace} & Adv-StyleGAN3 & 0.9674& 0.7251 & 0.8490 & 0.3535\\
        & & MSE-StyleGAN3 & 0.9773 & 0.8773 & 0.8954 & 0.6275\\
        &  & CLIP-FTI & 0.9619 & 0.7484 & 0.8397 & 0.3904\\
        &  & LBFTI & \textbf{0.9984} & \textbf{0.9837} & \textbf{0.9937} & \textbf{0.9408} \\
        \midrule
        \multirow{8}{*}{ElasticFace}
        & \multirow{4}{*}{ElasticFace} & Adv-StyleGAN3 & 0.9889 & 0.7210 & 0.9339 & 0.4113 \\
        & & MSE-StyleGAN3 & 0.9932 & 0.8250 & 0.9590 & 0.5500\\
        &  & CLIP-FTI & 0.9961 & 0.8662& 0.9680 & 0.6139 \\
        & & LBFTI & \textbf{0.9984} & \textbf{0.9837} & \textbf{0.9871} & \textbf{0.7434} \\
        \cmidrule(lr){2-7}
        & \multirow{4}{*}{ArcFace} & Adv-StyleGAN3 & 0.9971 & 0.7552& 0.9412 & 0.4576 \\
        & & MSE-StyleGAN3 & 0.9756 & 0.7364 & 0.8999 & 0.4696\\
        & & CLIP-FTI & 0.9892 & 0.9011 & 0.9423 & 0.6872\\
        &  & LBFTI & \textbf{0.9986} & \textbf{0.9226} & \textbf{0.9587} & \textbf{0.7412} \\
        \bottomrule
    \end{tabular}
    }
\end{table}

\begin{table}[h]
    \centering
    \small 
    \caption{TARs on the AgeDB dataset.}
    \label{tab:tar_agedb}
    \resizebox{1.0\linewidth}{!}{
    \begin{tabular}{@{}c|c|c|*{3}{>{\centering\arraybackslash}p{1.1cm}|}>{\centering\arraybackslash}p{1.1cm}@{}}
        \toprule
        \textbf{\multirow{2}{*}{\makecell{Databases \\and Losses}}} & \textbf{\multirow{2}{*}{\makecell{Target Models/ \\Unseen Models}}} & \textbf{Inverters} & \multicolumn{2}{c|}{\textbf{TAR$\uparrow$ at 1\%FAR}} & \multicolumn{2}{c@{}}{\textbf{TAR$\uparrow$ at 0.1\%FAR}} \\
        \cmidrule(lr){4-5} \cmidrule(lr){6-7}
        & & & \textbf{Type-I} & \textbf{Type-II} & \textbf{Type-I} & \textbf{Type-II} \\
        \midrule
        \multirow{8}{*}{ArcFace}
        & \multirow{4}{*}{ArcFace} & Adv-StyleGAN3 & 0.9978 & 0.4379 & 0.9868 & 0.2385 \\
        & & MSE-StyleGAN3 & 0.9970 & 0.4337 & 0.9750 & 0.2383\\
        & & CLIP-FTI & \textbf{0.9982} & 0.4080 & 0.9806 & 0.2102 \\
        & & LBFTI & 0.9965 & \textbf{0.5268} & \textbf{0.9907} & \textbf{0.3299} \\
        \cmidrule(lr){2-7}
        & \multirow{4}{*}{ElasticFace} & Adv-StyleGAN3 & 0.7977 & 0.2898 & 0.5856 & 0.1255\\
        & & MSE-StyleGAN3 & 0.8344 & 0.3280 & 0.6531 & 0.1503\\
        &  & CLIP-FTI & 0.8045 & 0.2951 & 0.5923 & 0.1284\\
        & & LBFTI & \textbf{0.9309} & \textbf{0.4461} & \textbf{0.8573} & \textbf{0.2488} \\
        \midrule
        \multirow{8}{*}{ElasticFace}
        & \multirow{4}{*}{ElasticFace} & Adv-StyleGAN3 & 0.9181 & 0.3141 & 0.8167 & 0.1338 \\
        & & MSE-StyleGAN3 & 0.9578 & 0.3125 & 0.8603 & 0.1324 \\
        &  & CLIP-FTI & 0.9302 & 0.3122 & 0.8567 & 0.1346 \\
        & & LBFTI & \textbf{0.9601} & \textbf{0.3222} & \textbf{0.8692} & \textbf{0.1346} \\
        \cmidrule(lr){2-7}
        & \multirow{4}{*}{ArcFace} & Adv-StyleGAN3 & 0.7892 & 0.3345 & 0.6397 & 0.1670 \\
        & & MSE-StyleGAN3 & 0.8099 & 0.3317 & 0.6580 & 0.1626 \\
        & & CLIP-FTI & 0.8058 & 0.3306 & 0.6434  & \textbf{0.1706} \\
        &  & LBFTI & \textbf{0.8180} & \textbf{0.3355} & \textbf{0.6614} & 0.1654 \\
        \bottomrule
    \end{tabular}
    }
\end{table}

\begin{figure}[h]
  \vskip 0.2in
  \begin{center}
    \centerline{\includegraphics[width=1.0\columnwidth]{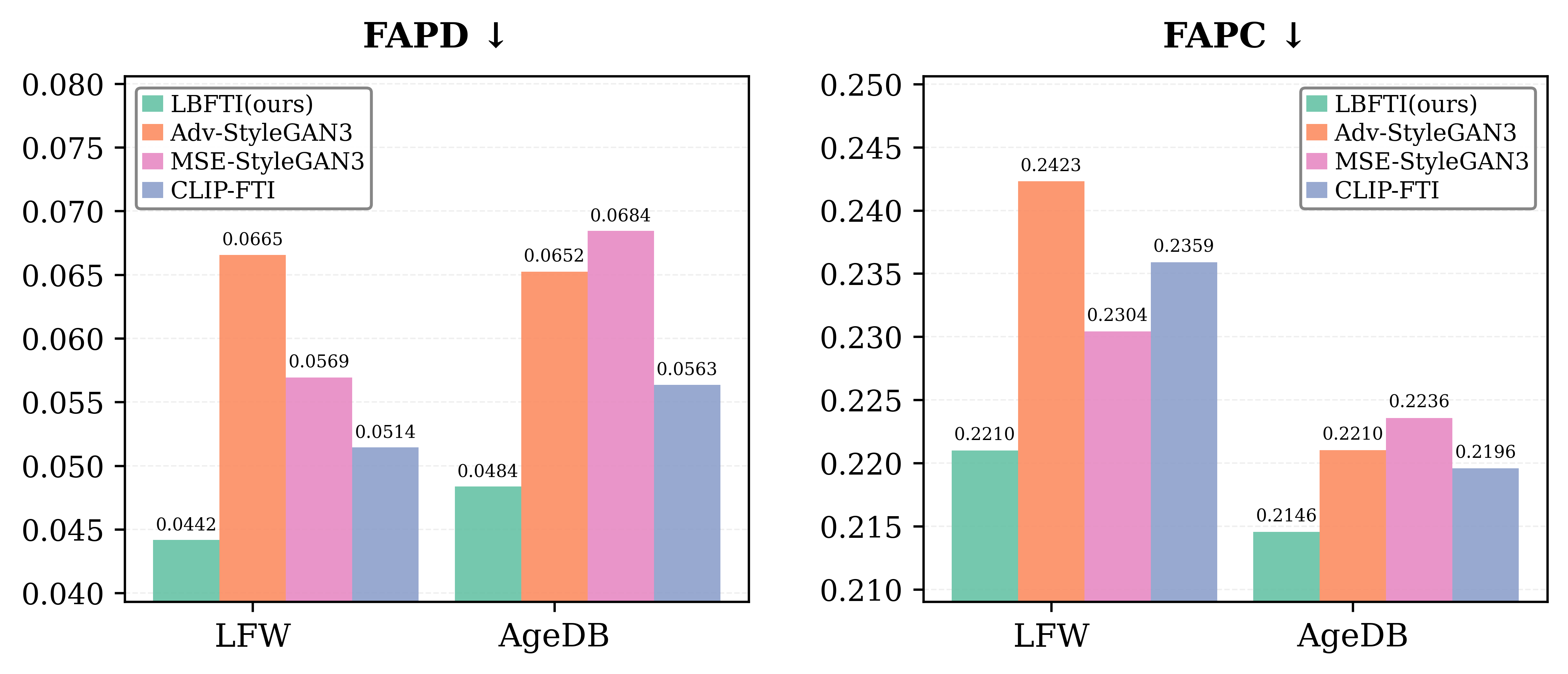}}
    \caption{Similarity of foreground layers between original and reconstructed face images on the LFW and AgeDB datasets with ArcFace as the target model.}
    \label{fig:fapdfapc}
  \end{center}
\end{figure}

\textbf{Performance on human-perceptual consistency}. Fig. \ref{fig:fapdfapc} shows the comparison results of FAPD and FAPC on the LFW and AgeDB datasets with ArcFace as the target model. It can be seen that both FAPD and FAPC metrics of LBFTI are significantly lower than those of the comparison methods, reaching 0.0442 and 0.2210 on the LFW dataset, which are 33.5\% and 8.8\% lower than those of Adv-StyleGAN3 (0.0665 and 0.2423), respectively. On the AgeDB dataset, LBFTI still maintains the lowest FAPD and FAPC, verifying its ability to restore facial feature details in complex scenarios.

\begin{figure}[h]
  \vskip 0.2in
  \begin{center}
    \centerline{\includegraphics[width=1.0\columnwidth]{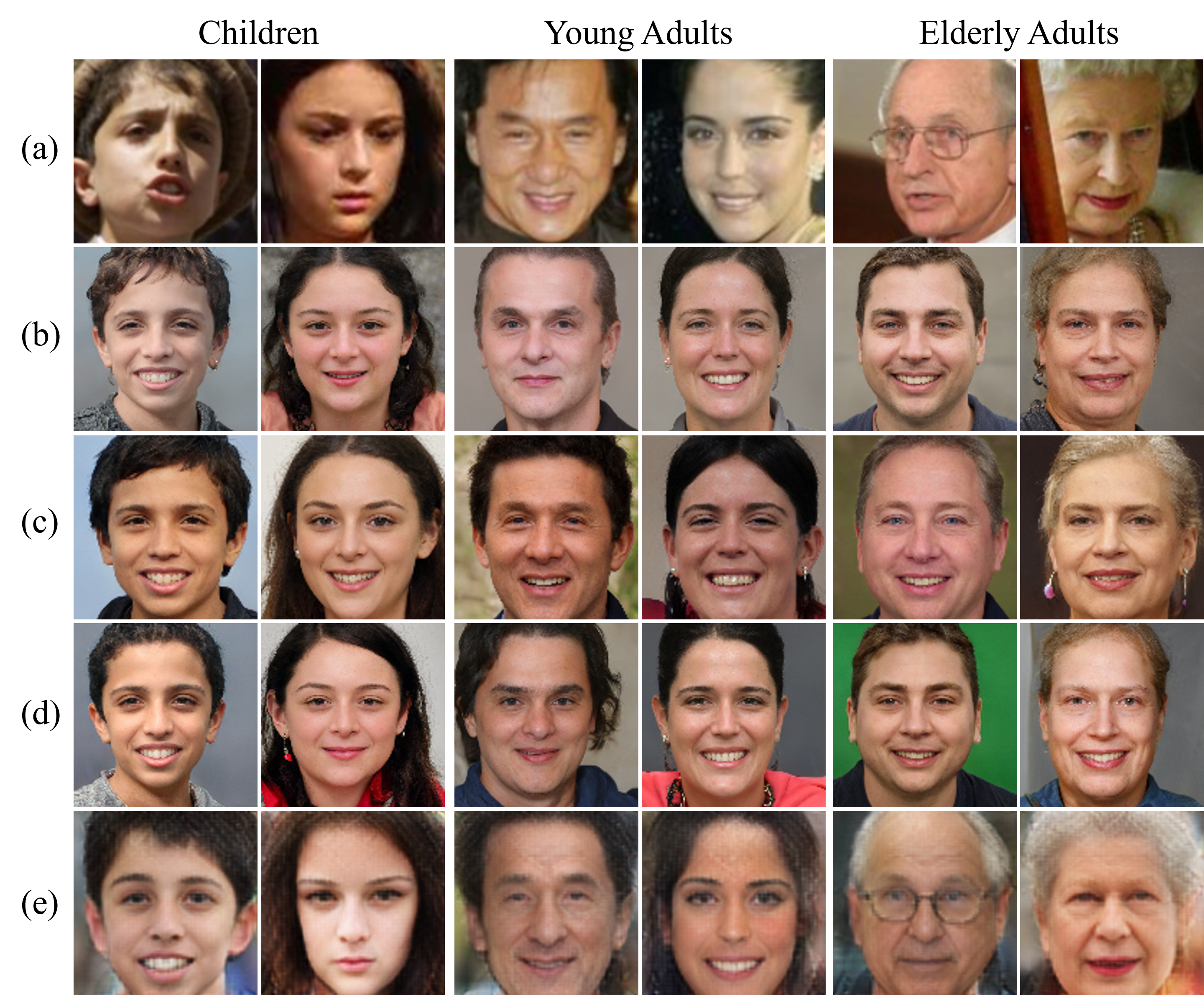}}
    \caption{Visual samples of original and reconstructed images on LFW: (a) original images, (b) reconstructed images by Adv-StyleGAN3, (c) reconstructed images by MSE-StyleGAN3, (d) reconstructed images by CLIP-FTI, and (e) reconstructed images by our LBFTI.}
    \label{fig:samples}
  \end{center}
\end{figure}

\textbf{Qualitative visualization}. Fig. \ref{fig:samples} shows partial qualitative results on the LFW dataset with ArcFace as the target model. It can be observed that images generated by Adv-StyleGAN3, MSE-StyleGAN3, and CLIP-FTI have problems such as inconsistent facial features (e.g., lack of eye details and distorted mouth shape), while images reconstructed by LBFTI are more similar to the original images in eyebrow texture, eye contour, nose structure, and mouth shape, with more realistic human visual perception.

\subsection{Ablation Experiments}

\begin{figure}[h]
  \vskip 0.2in
  \begin{center}
    \centerline{\includegraphics[width=1.0\columnwidth]{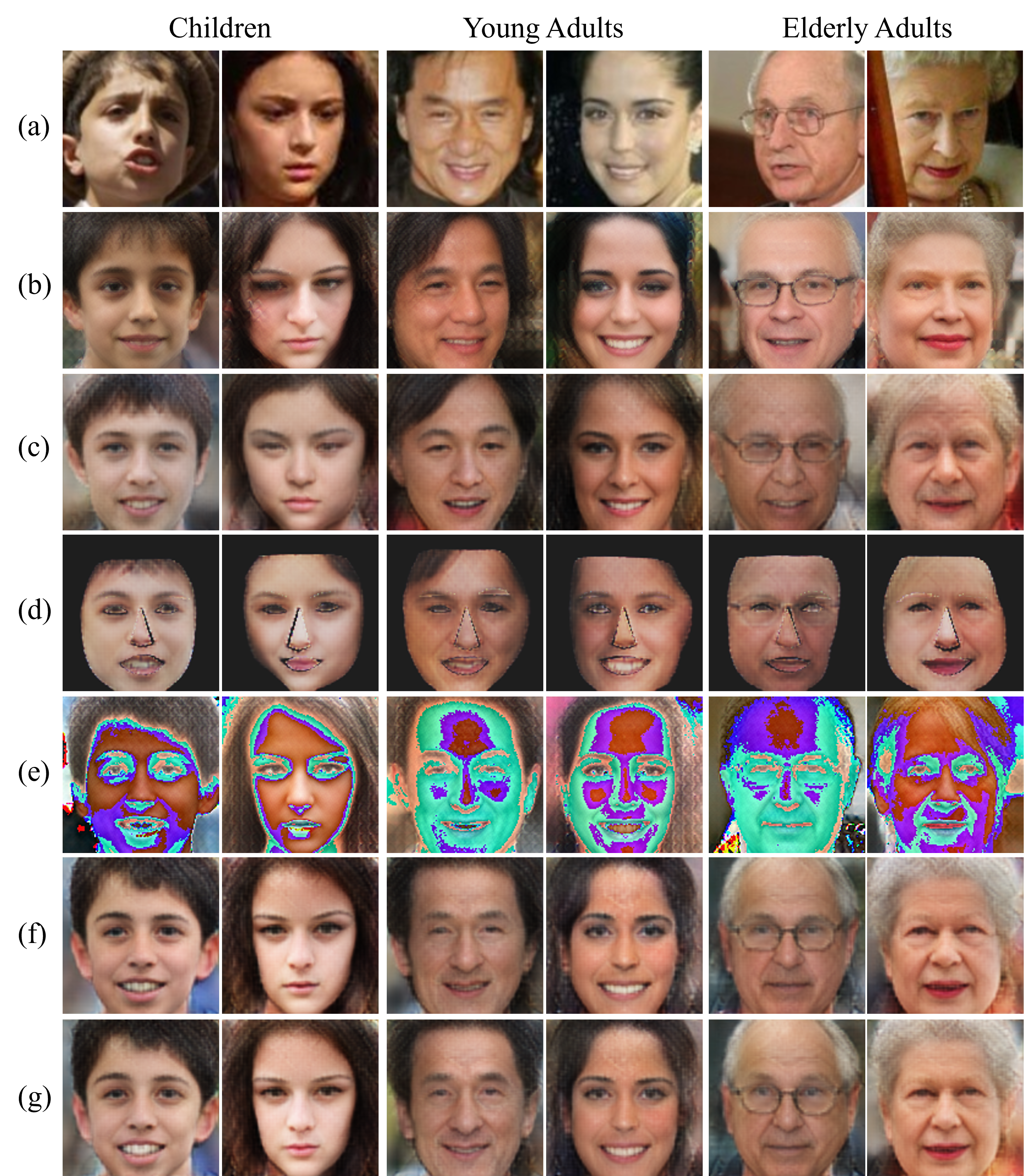}}
    \caption{Visual samples of original and images reconstructed by LBFTI on LFW: (a) original images, (b)-(g) reconstructed samples corresponding to the six ablation settings listed in rows 1 to 6 of Table \ref{tab:ablation_lfw}.}

    \label{fig:ablation_lfw}
  \end{center}
\end{figure}

To quantitatively validate the contribution of each core component of LBFTI, we conduct five groups of ablation experiments on the LFW dataset with ArcFace as the target face recognition model. As summarized in Table \ref{tab:ablation_lfw}, the first five rows correspond to four ablated model variants, while the sixth row shows the performance of the full LBFTI baseline model. Fig. \ref{fig:ablation_lfw} shows partial qualitative results on the LFW dataset with ArcFace as the target model. A detailed analysis of these ablation outcomes is presented in the following sections. Furthermore, the ablation experiments evaluated on the LFW dataset with ElasticFace as the target model are provided in \textbf{S.M.} C, which also demonstrates the necessity of each core component of our LBFTI.

\begin{table}[h] 
    \centering
    \small 
    \caption{Results of ablation experiments.} 
    \label{tab:ablation_lfw} 
    \resizebox{1.0\linewidth}{!}{
    
        \begin{tabular}{@{}>{\centering\arraybackslash}p{0.4cm} >{\centering\arraybackslash}p{0.4cm} >{\centering\arraybackslash}p{0.4cm} >{\centering\arraybackslash}p{0.4cm} >{\centering\arraybackslash}p{0.4cm}|c|c|c|c|c|c@{}}

        \toprule
        
        \textbf{\multirow{2}{*}{F-S1}} & \textbf{\multirow{2}{*}{M-S1}} & \textbf{\multirow{2}{*}{S2}} & \textbf{\multirow{2}{*}{FT-S2}} & \textbf{\multirow{2}{*}{S3}} & \multicolumn{2}{c|}{\textbf{TAR $\uparrow$ at 1\%FAR}} & \multicolumn{2}{c|}{\textbf{TAR$\uparrow$ at 0.1\%FAR}} & \textbf{\multirow{2}{*}{FAPD}}& \textbf{\multirow{2}{*}{FAPC}} \\
        \cmidrule(lr){6-7} \cmidrule(lr){8-9}
        &  &  &  & & \textbf{Type-I} & \textbf{Type-II} & \textbf{Type-I} & \textbf{Type-II} & & \\
        \midrule
        
        $\times$ & $\times$ & $\checkmark$ & $\checkmark$ & $\times$ & 0.9601 & 0.3222 & 0.8691 & 0.1359 & 0.0522 & 0.2277 \\
        $\checkmark$ & $\times$ & $\checkmark$ & $\checkmark$ & $\checkmark$ & 0.9995 & 0.9633 & 0.9986 & 0.8833 & 0.0520 & 0.2304 \\
        $\checkmark$ & $\checkmark$ & $\times$ & $\times$ & $\times$ & 0.6992 & 0.4254 & 0.4315 & 0.1557 & 0.0491 & 0.2450 \\
        $\checkmark$ & $\checkmark$ & $\checkmark$ & $\times$ & $\checkmark$ & 0.6049 & 0.3216 & 0.3530 & 0.1441 & 0.0510 & 0.2310 \\
        $\checkmark$ & $\checkmark$ & $\checkmark$ & $\checkmark$ & $\times$ & 0.9993 & 0.9804 & 0.9977 & 0.9245 & 0.0433 & 0.2215 \\
        $\checkmark$ & $\checkmark$ & $\checkmark$ & $\checkmark$ & $\checkmark$ & \textbf{0.9998} & \textbf{0.9861} & \textbf{0.9992} & \textbf{0.9509} & \textbf{0.0442} & \textbf{0.2210} \\
        \bottomrule
    \end{tabular}
    }
    \begin{tablenotes}[flushleft]
    \scriptsize 
    \item[†] Note: F-S1 denotes the ablation experiments on the foreground generator; M-S1 denotes the ablation experiments on the midground generator; S2 denotes the ablation experiments on the panorama generator; FT-S2 denotes the ablation experiments on whether the secondary injection of the facial template is implemented in Stage 2; S3 denotes the ablation experiments on the joint fine-tuning of all modules.
    \end{tablenotes}
\end{table}

\textbf{Necessity of layer-based generation}. Row 1 in Table \ref{tab:ablation_lfw} is obtained by ablating Stage 1 (and thus omitting Stage 3, the joint fine-tuning phase). Specifically, this ablation removes the foreground generators, midground generator, and the EncoderBlocks of the panorama generator from the LBFTI architecture. Experimental results show that at 0.1\% FAR, the Type-I TAR drops by more than 0.1, while the Type-II TAR plummets to about 0.1. This demonstrates that the layer-based generation paradigm is indispensable for preserving fine-grained facial attribute details and maintaining high identity consistency.

\textbf{Effectiveness of the midground generator}. Row 2 in Table \ref{tab:ablation_lfw} corresponds to the ablation of the midground generator in Stage 1, where foreground components are directly fused with the background via the panorama generator. This removal causes clear performance degradation: the Type-II TAR at 0.1\% FAR drops from 0.9509 to 0.8833, while both FAPD and FAPC worsen noticeably. These declines reveal that discarding the midground generator impairs visual fidelity and cross-layer harmony. This confirms the midground generator bridges foreground facial details and background context, effectively improving layer integration and enhancing both naturalness and identity consistency.

\textbf{Effectiveness of the panorama generator}. Row 3 in Table \ref{tab:ablation_lfw} is obtained by ablating Stage 2 (and consequently Stage 3). To synthesize complete face images under this setting, we perform naive pixel-level superposition on the foreground (eyebrows, eyes, nose, and mouth) and midground (skin) layers in Stage 1, resulting in reconstructed faces lacking background details and inter-regional coordination. The results indicate that Type-I TAR decreases by over 0.3 and Type-II TAR drops by more than 0.5. This verifies that standalone foreground and midground regions are insufficient to form a natural, authentic face image, and the panorama generator plays a pivotal role in supplementing background details and optimizing the overall coherence of reconstructed faces.

\textbf{Value of secondary template injection}. Row 4 in Table \ref{tab:ablation_lfw} corresponds to the model with the secondary facial template injection module in Stage 2 ablated. Experimental results show that without the supplementary injection of template features in Stage 2, the Type-II TAR at a 0.1\% FAR threshold drops drastically to 0.1441. This indicates that layer-based reconstruction in Stage 1 may cause the loss of fine-grained identity-related features, and secondary template injection can effectively compensate for such feature loss, thereby significantly boosting the machine authentication pass rate of reconstructed images.

\textbf{Gain of joint fine-tuning}. Row 5 in Table \ref{tab:ablation_lfw} is obtained by ablating Stage 3. Results show that the Type-II TAR at a 0.1\% FAR threshold decreases from 0.9509 to 0.9245, accompanied by a slight elevation in FAPD and FAPC metrics. This indicates that joint fine-tuning can further optimize inter-module coordination and cross-region consistency of the model, thus enhancing the overall quality of reconstructed face images.

\section{Conclusion}

This paper presents a novel layer-based facial template inversion (LBFTI) method for identity-preserving fine-grained face reconstruction. By decomposing face images into foreground (eyebrows, eyes, nose, and mouth), midground (skin), and background (other parts) layers, LBFTI achieves fine-grained reconstruction of each layer using dedicated layer-specific generators and a rigorous three-stage training strategy. Experimental results show that LBFTI outperforms SOTA methods in both performance on machine authentication and human-perceptual consistency. Quantitatively, our LBFTI achieves superior TAR values, along with promising performance on FAPD, FAPC, and subjective questionnaire evaluations. These results demonstrate the advantages of our layer-based design, which could be further extended to various other generation tasks.

\bibliographystyle{ACM-Reference-Format}
\bibliography{main.bib}

\newpage

\appendix

\section{Questionnaire Survey}
\label{appendix:A}
\subsection{The Content of Questionnaire}

\begin{figure}[htbp]
    \centering
    \includegraphics[width=1.0\columnwidth]{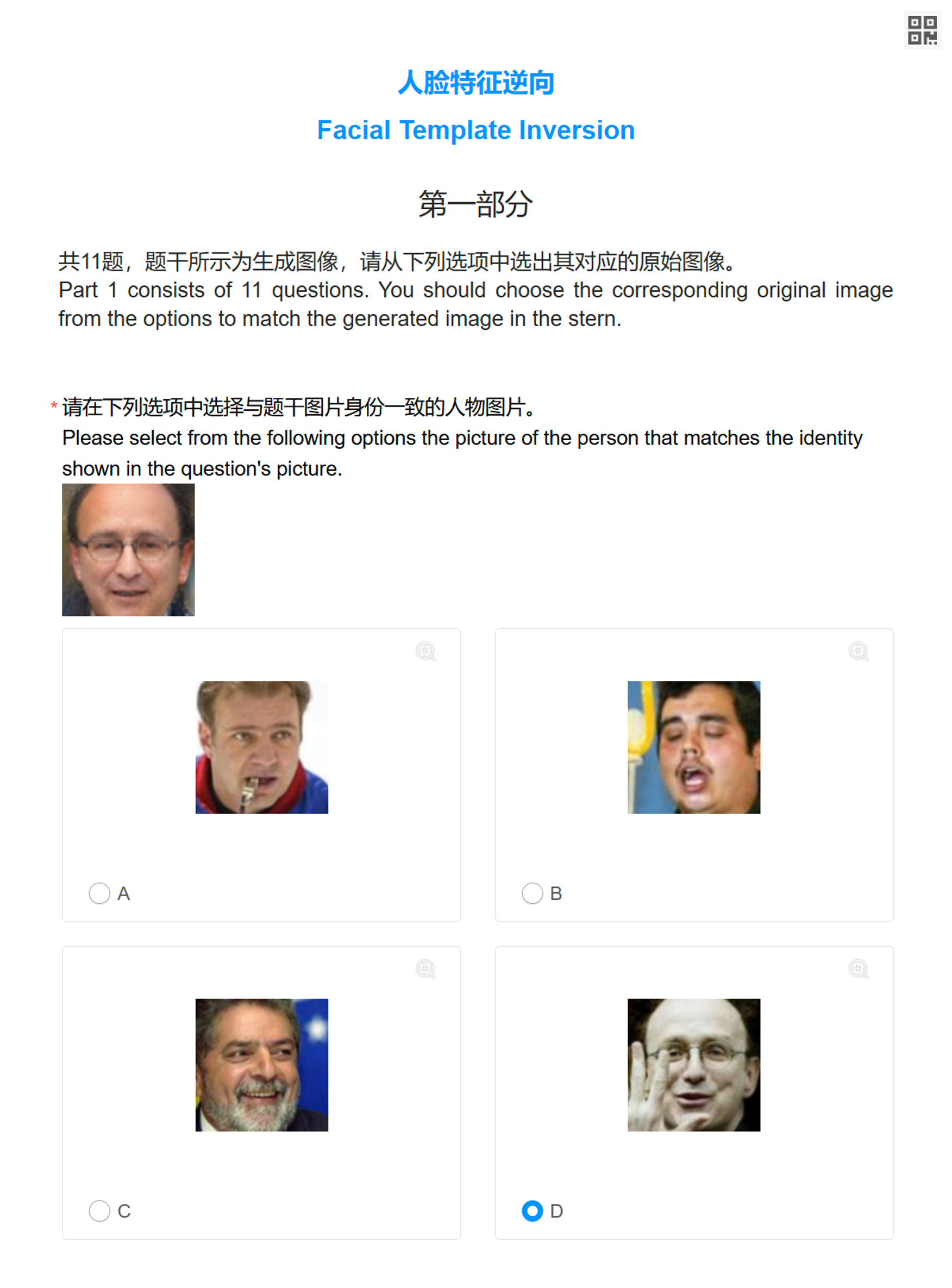}
    \caption{Illustration of Part \#1 in the questionnaire survey.}
    \label{fig:content_q1}
\end{figure}

\begin{figure}[htbp]
    \centering
    \includegraphics[width=1.0\columnwidth]{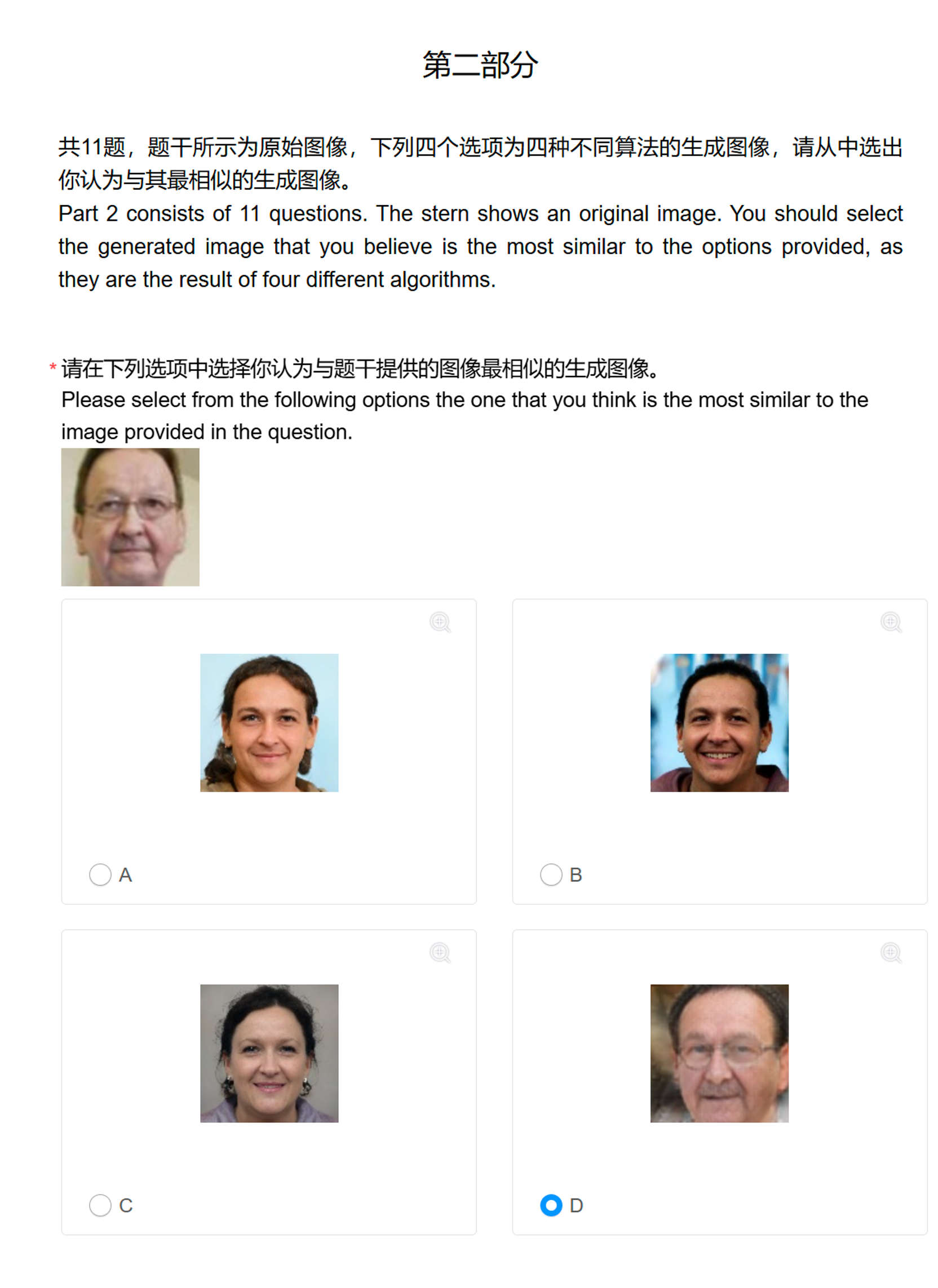}
    \caption{Illustration of Part \#2 in the questionnaire survey.}
    \label{fig:content_q2}
\end{figure}

\begin{figure}[htbp]
    \centering
    \includegraphics[width=1.0\columnwidth]{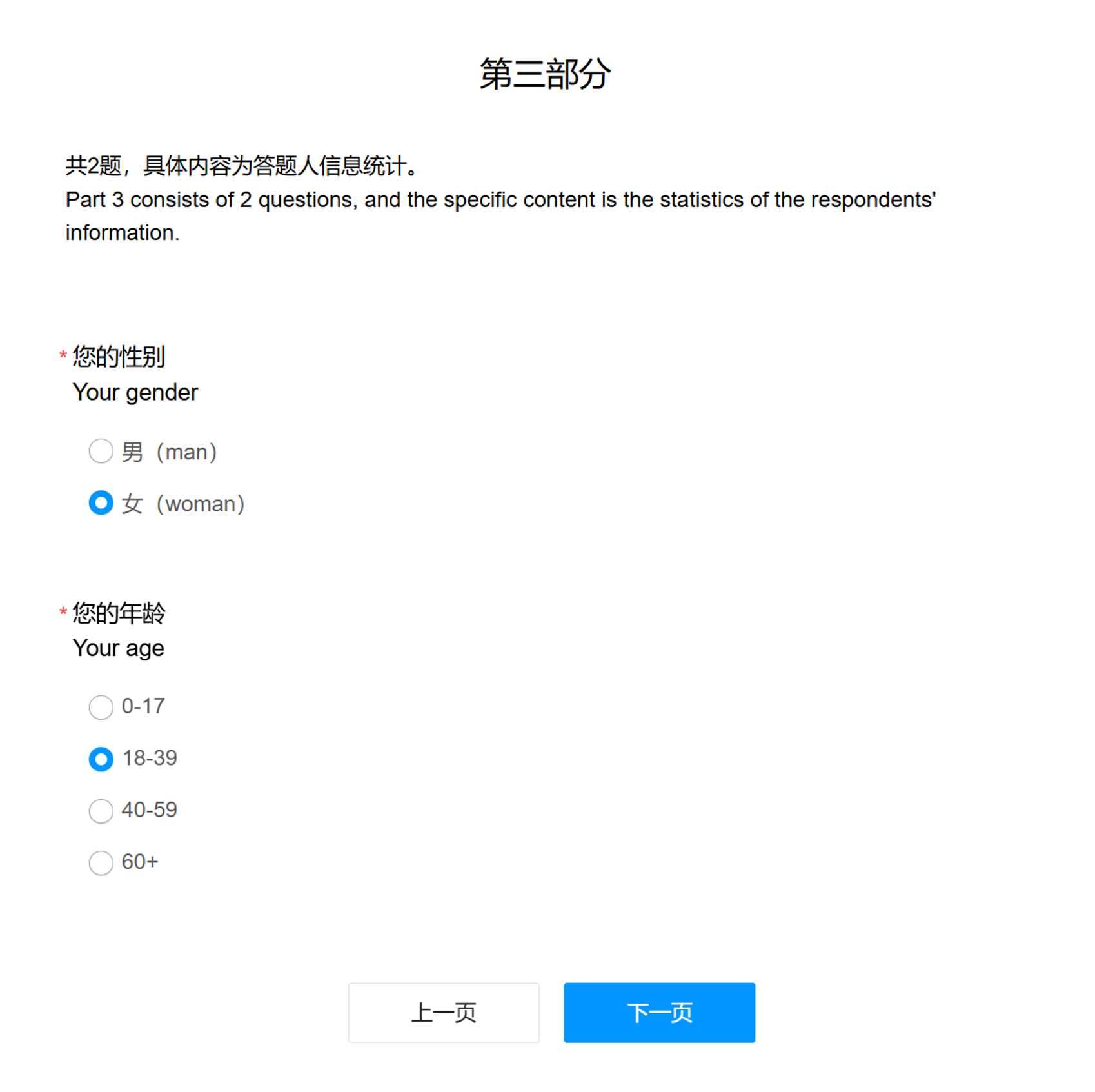}
    \caption{Illustration of Part \#3 in the questionnaire survey.}
    \label{fig:content_q3}
\end{figure}

To evaluate human perceptual recognition of reconstructed face images, we conduct a subjective questionnaire survey where participants visually match generated images with original ones. The questionnaire contains 24 single-choice questions divided into three sections, as illustrated in Figs. \ref{fig:content_q1}, \ref{fig:content_q2}, and \ref{fig:content_q3}.

\begin{itemize}
    \item \textbf{Part \#1 (11 questions)}. Participants are presented with face images reconstructed by our LBFTI on the LFW dataset and asked to select the corresponding original image from four candidate options. Apart from the correct original sample, the other three distractors are randomly selected from the LFW dataset.
    \item \textbf{Part \#2 (11 questions)}. Given an original LFW face image, volunteers are required to choose the most visually similar reconstruction among four images generated by different methods: our LBFTI, Adv-StyleGAN3, MSE-StyleGAN3, and CLIP-FTI, as introduced in Sec. \ref{sec:compare}.
    \item \textbf{Part \#3 (2 questions)}. Two additional questions collect participants’ gender and age information. This enables further analysis of identity recognition performance across different demographic groups.
\end{itemize}

In addition, we embed a validation item with an obvious correct answer in both Part \#1 and Part \#2 to filter out careless and invalid responses. The questionnaire is distributed online via the link (\url{https://ks.wjx.com/vm/rCyczwU.aspx\#}) and shared through WeChat and other social platforms. In total, 200 responses are collected, and 156 valid questionnaires are retained for subsequent analysis.

\subsection{The Results of Questionnaire Survey}

\begin{figure}[h]
  \vskip 0.2in
  \centering
  \includegraphics[width=1.0\columnwidth]{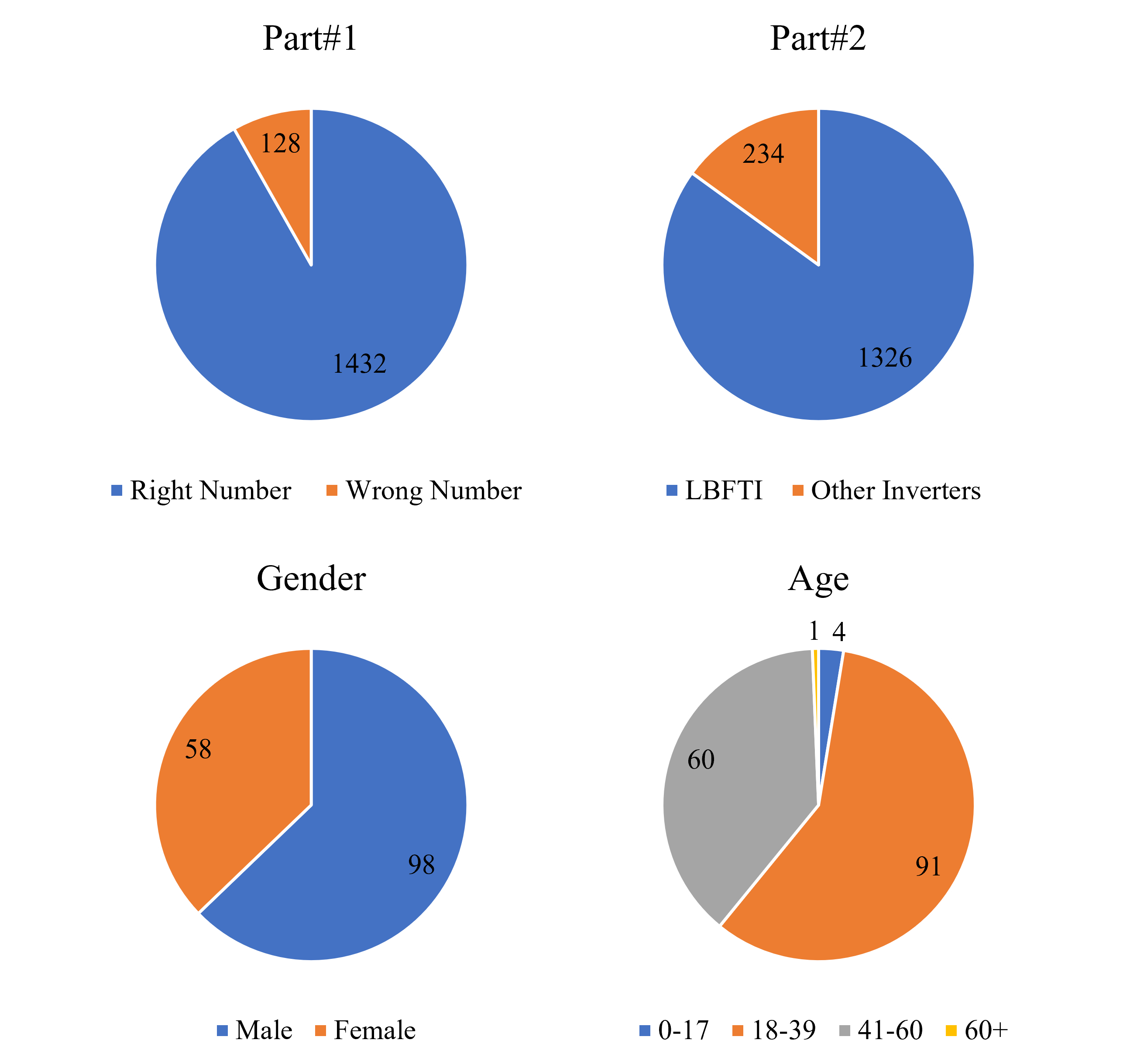}
  \caption{Statistical results of the questionnaire survey.}
  \label{fig:result_q}
\end{figure}

The statistical outcomes of the questionnaire study are summarized in Fig. \ref{fig:result_q}, and several key findings are presented as follows:

\begin{itemize}
    \item Based on the results of Part \#1, face images reconstructed by LBFTI achieve a human recognition accuracy of 91.79\%. Demographic comparison reveals subtle performance differences across groups: female participants reach an accuracy of 97.93\%, while male participants obtain 88.26\%. Young participants achieve an accuracy of 97.50\%, and middle-aged participants reach 82.67\%.
    \item The results of Part \#2 demonstrate that LBFTI outperforms the three state-of-the-art methods in visual reconstruction quality. The reconstructions generated by our LBFTI obtain 85.00\% of the preference votes, which significantly exceeds the total votes of all baseline approaches. Across gender and age groups, female participants achieve a preference rate of 95.86\%, while male participants reach 78.57\%. Young participants obtain a preference rate of 94.58\%, and middle-aged participants reach 69.67\%.
\end{itemize}

\section{The Complete Cross-model Generalization Performance of LBFTI}
\label{appendix:B}

 Tables \ref{tab:tar_lfw_full} and \ref{tab:tar_agedb_full} present the full TAR results on the LFW and AgeDB datasets, respectively, to comprehensively evaluate the cross-model generalization capability of our proposed LBFTI framework across seven mainstream face recognition backbones, namely ArcFace, ElasticFace, MagFace, FaceNet, AdaFace, SphereFace, and CurricularFace.

 As observed from the results on the LFW dataset, which serves as a widely adopted benchmark in face recognition research. When ArcFace is adopted as the target feature extractor, the Type-II TAR of our LBFTI reaches 0.9509 at 0.1\% FAR under matched ArcFace verification. When generalized to the unseen ElasticFace model, the Type-II TAR remains at a high level of 0.9408. Across all seven evaluated face recognition methods, including MagFace, CurricularFace, SphereFace, AdaFace, and FaceNet, LBFTI demonstrates consistently stable and outstanding performance. Specifically, at 1\% FAR, all models maintain a Type-I TAR above 90\%, while most Type-II TAR values exceed 80\%. Even at the stricter 0.1\% FAR, the overall performance undergoes only a slight decline: most Type-I TARs stay above 80\%, and nearly all Type-II TARs remain higher than 60\%. Such consistent performance across different evaluation protocols and diverse recognition architectures strongly validates the excellent cross-model generalization robustness of our proposed LBFTI framework.
 
 On the more challenging AgeDB dataset featuring dramatic age variations that raise the difficulty of facial template inversion, our LBFTI still maintains promising performance. Significant age-induced facial alterations greatly increase the complexity of inversion tasks on this dataset. Even under such challenging settings, LBFTI achieves remarkable results across all evaluated backbones. At 1\% FAR, the Type-I TAR of most tested models exceeds 75\%, with a large proportion further reaching over 90\%. Under the stricter 0.1\% FAR, the Type-I TAR for most models remains consistently above 60\%, and several cases even surpass 85\%. These results demonstrate that our LBFTI can effectively capture fine-grained identity features. It maintains stable cross-model generalization and reliable identity matching capability, achieving robust performance against severe age variations and complex real-world disturbances.

\begin{table}[h]
    \centering
    \small 
    \caption{TARs on the LFW dataset.}
    \label{tab:tar_lfw_full}
    \resizebox{1.0\linewidth}{!}{
    \begin{tabular}{@{}c|c|*{4}{>{\centering\arraybackslash}p{1.1cm}}@{}}
        \toprule
        \textbf{\multirow{2}{*}{\makecell{Databases \\and Losses}}} & \textbf{\multirow{2}{*}{\makecell{Target Models/ \\Unseen Models}}} & \multicolumn{2}{c|}{\textbf{TAR$\uparrow$ at 1\%FAR}} & \multicolumn{2}{c@{}}{\textbf{TAR$\uparrow$ at 0.1\%FAR}} \\
        \cmidrule(lr){3-4} \cmidrule(lr){5-6}
        & & \textbf{Type-I} & \textbf{Type-II} & \textbf{Type-I} & \textbf{Type-II} \\
        \midrule
        \multirow{7}{*}{ArcFace}
        & ArcFace & 0.9998 & 0.9861 & 0.9992 & 0.9509 \\
        & ElasticFace & 0.9984 & 0.9837 & 0.9937 & 0.9408 \\
        & MagFace & 0.9969 & 0.9730 & 0.9826 & 0.8945 \\
        & FaceNet & 0.9314 & 0.8694 & 0.8065 & 0.5954 \\
        & AdaFace & 0.9510 & 0.9100 & 0.8595 & 0.7482 \\
        & SphereFace & 0.9941 & 0.9635 & 0.9661 & 0.8537 \\
        & CurricularFace & 0.9957 & 0.9738 & 0.9779 & 0.9094 \\
        \cmidrule(lr){1-6}
        \multirow{7}{*}{ElasticFace}
        & ElasticFace & 0.9984 & 0.9374 & 0.9871 & 0.7434 \\
         & ArcFace & 0.9886 & 0.8226 & 0.9587 & 0.7412 \\
        & MagFace & 0.9845 & 0.9408 & 0.9480 & 0.7798 \\
        & FaceNet & 0.9413 & 0.7967 & 0.8301 & 0.4535 \\
        & AdaFace & 0.9488 & 0.8710 & 0.8489 & 0.6354 \\
        & SphereFace & 0.9851 & 0.9294 & 0.9398 & 0.7131 \\
        & CurricularFace & 0.9832 & 0.9477 & 0.9476 & 0.8087 \\
        \bottomrule
    \end{tabular}
    }
\end{table}

\begin{table}[h]
    \centering
    \small 
    \caption{TARs on the AgeDB dataset.}
    \label{tab:tar_agedb_full}
    \resizebox{1.0\linewidth}{!}{
    \begin{tabular}{@{}c|c|*{4}{>{\centering\arraybackslash}p{1.1cm}}@{}}
        \toprule
        \textbf{\multirow{2}{*}{\makecell{Databases \\and Losses}}} & \textbf{\multirow{2}{*}{\makecell{Target Models/ \\Unseen Models}}} & \multicolumn{2}{c|}{\textbf{TAR$\uparrow$ at 1\%FAR}} & \multicolumn{2}{c@{}}{\textbf{TAR$\uparrow$ at 0.1\%FAR}} \\
        \cmidrule(lr){3-4} \cmidrule(lr){5-6}
        & & \textbf{Type-I} & \textbf{Type-II} & \textbf{Type-I} & \textbf{Type-II} \\
        \midrule
        \multirow{7}{*}{ArcFace}
        & ArcFace & 0.9965 & 0.5268 & 0.9907 & 0.3299 \\
        & ElasticFace & 0.9309 & 0.4461 & 0.8573 & 0.2488 \\
        & MagFace & 0.9443 & 0.4404 & 0.8556 & 0.2423 \\
        & FaceNet & 0.7651 & 0.3459 & 0.5698 & 0.1433 \\
        & AdaFace & 0.6786 & 0.2179 & 0.4588 & 0.0832 \\
        & SphereFace & 0.9258 & 0.4747 & 0.8154 & 0.2555 \\
        & CurricularFace & 0.9064 & 0.4051 & 0.8022 & 0.2247 \\
        \cmidrule(lr){1-6}
        \multirow{7}{*}{ElasticFace}
        & ElasticFace & 0.9601 & 0.3222 & 0.8692 & 0.1356 \\
        & ArcFace & 0.8180 & 0.3355 & 0.6614 & 0.1654 \\
        & MagFace & 0.7805 & 0.3335 & 0.6088 & 0.1639 \\
        & FaceNet & 0.6572 & 0.2865 & 0.4502 & 0.1089 \\
        & AdaFace & 0.6434 & 0.1690 & 0.3971 & 0.0496 \\
        & SphereFace & 0.7955 & 0.3675 & 0.6100 & 0.1670 \\
        & CurricularFace & 0.7662 & 0.3112 & 0.5964 & 0.1533 \\
        \bottomrule
    \end{tabular}
    }
\end{table}

\section{Ablation Experiment Results on the LFW Dataset with ElasticFace as the Target Model}
\label{appendix:C}

\begin{figure}[h]
  \vskip 0.2in
  \begin{center}
    \centerline{\includegraphics[width=1.0\columnwidth]{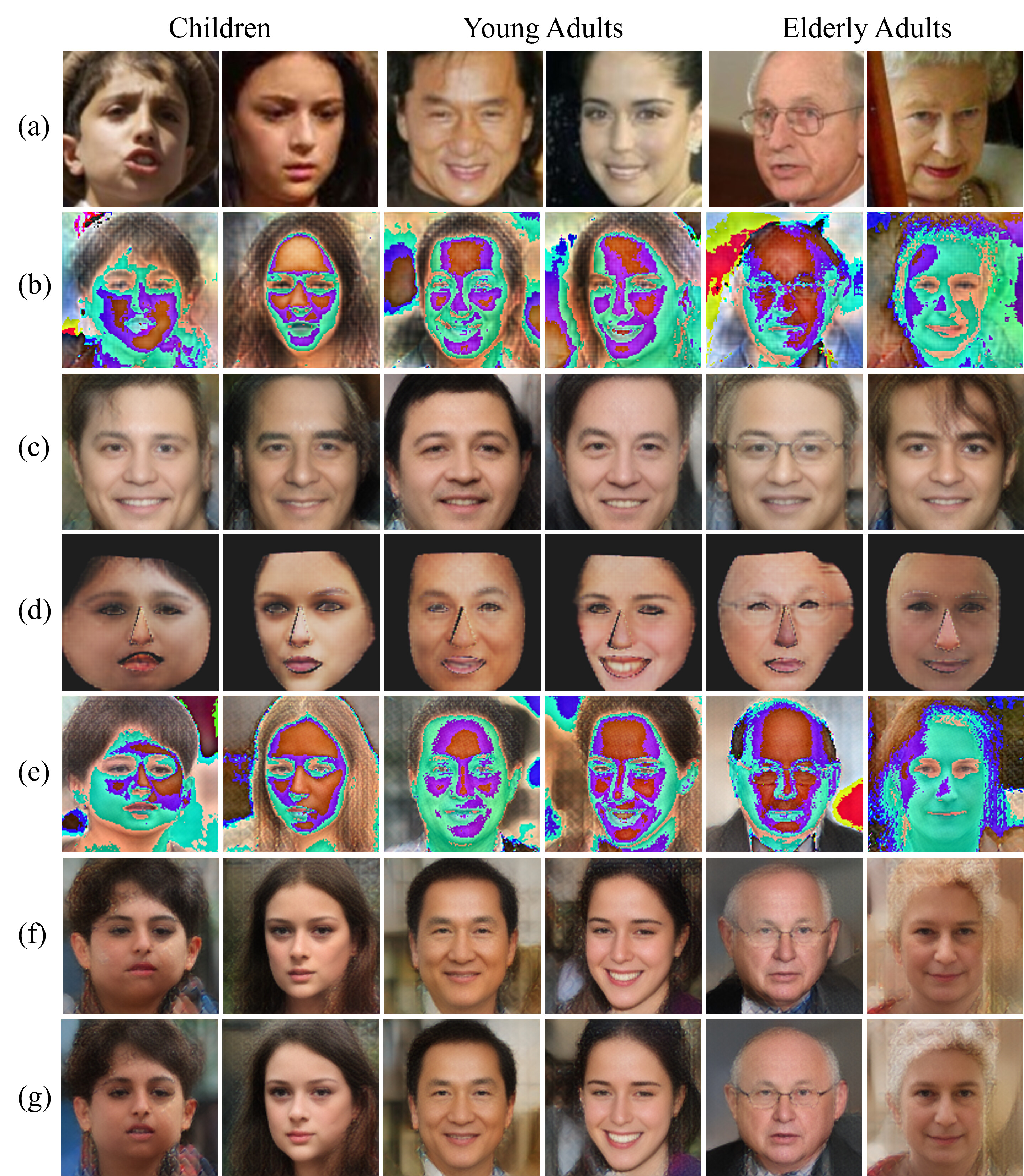}}
    \caption{Visual samples of original and images reconstructed by LBFTI on LFW: (a) original images, (b)-(g) reconstructed samples corresponding to the six ablation settings listed in rows 1 to 6 of Table \ref{tab:ablation_lfw_ela}.}

    \label{fig:ablation_lfw_ela}
  \end{center}
\end{figure}

 The main manuscript has presented the ablation results on the LFW dataset with ArcFace adopted as the target recognition model. For supplementary validation, this section further supplements the experimental outcomes using ElasticFace as the target model. To quantitatively verify the contribution of each core component within the LBFTI framework, we conduct five groups of ablation experiments on the LFW dataset with ElasticFace serving as the target face recognition model. As summarized in Table \ref{tab:ablation_lfw_ela}, the first five rows correspond to different ablated variants, while the sixth row reports the performance of the complete LBFTI baseline model. Fig. \ref{fig:ablation_lfw_ela} provides representative qualitative visualization results of these ablation experiments under the ElasticFace setting on the LFW dataset. 

 Notably, the conclusions observed in this section are highly consistent with those obtained from the ArcFace experiments in the main text. The performance degradation of each ablated variant clearly demonstrates that every core component of LBFTI is indispensable and rationally designed. Only when all modules work collaboratively can the framework achieve its optimal overall performance, further verifying the rationality and effectiveness of our architectural design across different target recognition models.

 \begin{table}[h] 
    \centering
    \small 
    \caption{Results of ablation experiments.} 
    \label{tab:ablation_lfw_ela} 
    \resizebox{1.0\linewidth}{!}{
    
        \begin{tabular}{@{}>{\centering\arraybackslash}p{0.4cm} >{\centering\arraybackslash}p{0.4cm} >{\centering\arraybackslash}p{0.4cm} >{\centering\arraybackslash}p{0.4cm} >{\centering\arraybackslash}p{0.4cm}|c|c|c|c|c|c@{}}

        \toprule
        
        \textbf{\multirow{2}{*}{F-S1}} & \textbf{\multirow{2}{*}{M-S1}} & \textbf{\multirow{2}{*}{S2}} & \textbf{\multirow{2}{*}{FT-S2}} & \textbf{\multirow{2}{*}{S3}} & \multicolumn{2}{c|}{\textbf{TAR $\uparrow$ at 1\%FAR}} & \multicolumn{2}{c|}{\textbf{TAR$\uparrow$ at 0.1\%FAR}} & \textbf{\multirow{2}{*}{FAPD}}& \textbf{\multirow{2}{*}{FAPC}} \\
        \cmidrule(lr){6-7} \cmidrule(lr){8-9}
        &  &  &  & & \textbf{Type-I} & \textbf{Type-II} & \textbf{Type-I} & \textbf{Type-II} & & \\
        \midrule
        
        $\times$ & $\times$ & $\checkmark$ & $\checkmark$ & $\times$ & 0.1087 & 0.0270 & 0.0387 & 0.0055 & 0.1244 & 0.5046 \\
        $\checkmark$ & $\times$ & $\checkmark$ & $\checkmark$ & $\checkmark$ & 0.0880 & 0.0028 & 0.0089 & 0.0001 & 0.0610 & 0.2224 \\
        $\checkmark$ & $\checkmark$ & $\times$ & $\times$ & $\times$ & 0.9036 & 0.4863 & 0.7229 & 0.2242 & 0.0457 & 0.2387 \\
        $\checkmark$ & $\checkmark$ & $\checkmark$ & $\times$ & $\checkmark$ & 0.0953 & 0.0231 & 0.0330 & 0.0037 & 0.0438 & 0.2197 \\
        $\checkmark$ & $\checkmark$ & $\checkmark$ & $\checkmark$ & $\times$ & 0.9963 & 0.8805 & 0.9780 & 0.6157 & 0.0416 & 0.2187 \\
        $\checkmark$ & $\checkmark$ & $\checkmark$ & $\checkmark$ & $\checkmark$ & \textbf{0.9984} & \textbf{0.9837} & \textbf{0.9871} & \textbf{0.7434} & \textbf{0.0417} & \textbf{0.2182} \\
        \bottomrule
    \end{tabular}
    }
    \begin{tablenotes}[flushleft]
    \scriptsize 
    \item[†] Note: F-S1 denotes the ablation experiments on the foreground generator; M-S1 denotes the ablation experiments on the midground generator; S2 denotes the ablation experiments on the panorama generator; FT-S2 denotes the ablation experiments on whether the secondary injection of the facial template is implemented in Stage 2; S3 denotes the ablation experiments on the joint fine-tuning of all modules.
    \end{tablenotes}
 \end{table}

\clearpage
\end{document}